\documentclass{article} 
\usepackage{arxiv,times}

\usepackage{amsmath,amsfonts,bm}

\def\eqref#1{equation~\ref{#1}}

\def\1{\bm{1}}

\DeclareMathAlphabet{\mathsfit}{\encodingdefault}{\sfdefault}{m}{sl}
\SetMathAlphabet{\mathsfit}{bold}{\encodingdefault}{\sfdefault}{bx}{n}

\usepackage{hyperref}
\usepackage{url}
\usepackage{xcolor}         
\usepackage{amsmath}
\usepackage{amssymb}
\usepackage{pgf-pie}
\usepackage{graphicx}
\usepackage{pgfplots}
\usetikzlibrary{patterns}
\pgfplotsset{compat=1.18}
\usepackage{graphicx}
\usepackage{pythonhighlight}
\usepackage{bbding}
\usepackage{booktabs} 

\usepackage{CJKutf8}

\newcommand{\datasetname}{M-LongDoc}

\newcommand{\datasize}{851}
\newcommand{\performanceincrease}{4.6\%}
\newcommand{\trainsize}{10,070}
\newcommand{\declarelogo}[0]{\includegraphics[height=.02\textwidth]{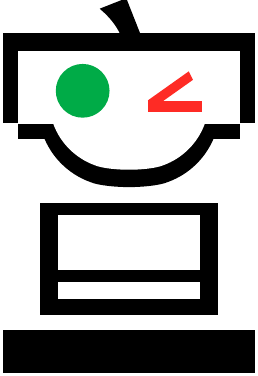}}

\title{\datasetname{}: A Benchmark and Retrieval-Aware Tuning for Long Multimodal Document Understanding}
\title{\datasetname{}: A Benchmark For Multimodal Super-Long Document Understanding And A Retrieval-Aware Tuning Framework}

\author{
\textbf{
Yew Ken Chia\thanks{Yew Ken and Chaoqun are students under the Joint PhD Program between Alibaba and their corresponding university. Work done while Liying was at Alibaba.
}
~\textsuperscript{\rm 1,~${\declarelogo}$}
\quad
Liying Cheng\textsuperscript{\rm ~1}
\quad Hou Pong Chan\footnotemark[2]\textsuperscript{\rm ~~~1}
\quad
Chaoqun Liu\textsuperscript{\rm ~1, ~2} 
} \\
\textbf{
~Maojia Song\textsuperscript{\rm ~${\declarelogo}$} 
\quad
Sharifah Mahani Aljunied\textsuperscript{\rm ~1} 
\quad
Soujanya Poria\textsuperscript{\rm ~${\declarelogo}$}
\quad 
Lidong Bing\thanks{~~Corresponding authors.}\textsuperscript{\rm ~~~1}
} \\
\\
\textsuperscript{\rm ${\declarelogo}$} Singapore University of Technology and Design ~~\\
\textsuperscript{\rm 1} DAMO Academy, Alibaba Group, Singapore~~  \\
\textsuperscript{\rm 2} Nanyang Technological University, Singapore \\
{\tt\{yewken\_chia, maojia\_song, sporia\}@mymail.sutd.edu.sg} \\
{\tt\{houpong.chan, chaoqun.liu, mahani.aljunied\}@alibaba-inc.com} \\
{\tt\{chengliying9581, binglidong\}@gmail.com}
}

\iclrfinalcopy 
\begin{document}

\maketitle

\begin{abstract}
The ability to understand and answer questions over documents can be useful in many business and practical applications. However, documents often contain lengthy and diverse multimodal contents such as texts, figures, and tables, which are very time-consuming for humans to read thoroughly. Hence, there is an urgent need to develop effective and automated methods to aid humans in this task. In this work, we introduce \datasetname{}, a benchmark of \datasize{} samples, and an automated framework to evaluate the performance of large multimodal models.
We further propose a retrieval-aware tuning approach for efficient and effective multimodal document reading. Compared to existing works, our benchmark consists of more recent and lengthy documents with hundreds of pages, while also requiring open-ended solutions and not just extractive answers. To our knowledge, our training framework is the first to directly address the retrieval setting for multimodal long documents. To enable tuning open-source models, we construct a training corpus in a fully automatic manner for the question-answering task over such documents. 
Experiments show that our tuning approach achieves a relative improvement of \performanceincrease{} for the correctness of model responses, compared to the baseline open-source models. Our data, code, and models are available at \url{https://multimodal-documents.github.io}
\end{abstract} 

\section{Introduction}

The ability to comprehend long and complex multi-modal documents and respond to user queries about them is crucial in various practical applications such as business intelligence analysis, academic literature review, and legal research~\citep{Mathew2020DocVQAAD}. 
Recently, large multimodal models such as GPT-4V \citep{gpt4vision} have shown great potential in processing and analyzing diverse types of information, including text, images, and even structured data \citep{charts-huang-etal-2024-lvlms}.
These models offer the promise of automating tasks that traditionally required extensive human effort, such as document analysis, information retrieval, and question-answering \citep{fujitake-2024-layoutllm}.
However, real-world documents often present significant challenges due to their length, complexity, and multimodal nature, containing a mix of text, figures, tables, and charts \citep{Faysse2024ColPaliED}.
Thus, it is not clear whether current models are capable of an in-depth understanding of lengthy multimodal documents. 
On the other hand, while existing benchmarks have fostered great progress in document understanding, they often fall short in representing these challenges, typically focusing on documents with less than 50 pages,
and limited to simpler extraction-based questions \citep{Ma2024MMLongBenchDocBL}. 

\begin{figure*}[t]
    \centering
    \begin{minipage}{0.48\textwidth}
        \centering
        \includegraphics[width=\textwidth]{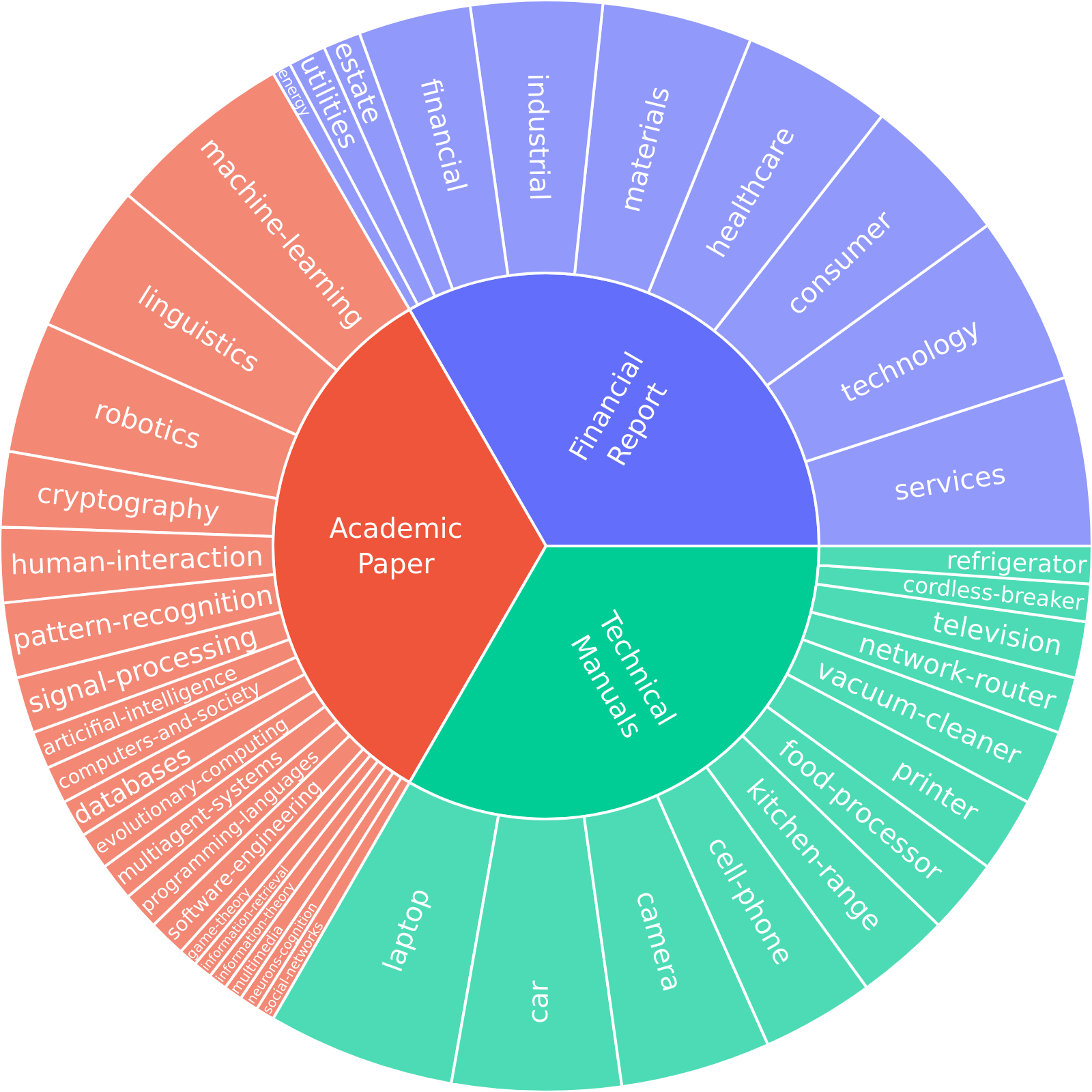}
        \caption{Data distribution of document topics 
        in our \datasetname{} benchmark.
        }
        \label{fig:data_pie}
    \end{minipage}%
    \hfill
    \begin{minipage}{0.48\textwidth}
        \centering
        \resizebox{\textwidth}{!}{
        \begin{tabular}{lccc}
        \toprule
         & Pages & Tokens & In-Depth \\
        \midrule
        DocVQA & 1.0 & 151.5 & \color{purple}{\XSolidBrush} \\
        ChartQA & 1.0 & 236.9 & \color{purple}{\XSolidBrush} \\
        InfoVQA & 1.2 & 288.0 & \color{purple}{\XSolidBrush} \\
        TAT-DQA & 1.1 & 577.0 & \color{purple}{\XSolidBrush} \\
        VisualWebBench & 1.0 & 452.4 & \color{purple}{\XSolidBrush} \\
        PWC & 12 & 7000 & \color{purple}{\XSolidBrush} \\
        MP-DocVQA & 8.3 & 2026.6 & \color{purple}{\XSolidBrush} \\
        DUDE & 5.7 & 1831.5 & \color{purple}{\XSolidBrush} \\
        SlideVQA & 20.0 & 151.5 & \color{purple}{\XSolidBrush} \\
        MMLongBench & 47.5 & 2030.5 & \color{purple}{\XSolidBrush} \\
        \textbf{Ours} & 210.8 & 120988.0 & \color{teal}{\Checkmark} \\
        \bottomrule
        \end{tabular}
        }
        \caption{
        Comparison of benchmarks along three dimensions: the number of pages per document, the number of tokens per document, and the nature of the responses required. Specifically, we assess whether each benchmark emphasizes in-depth, comprehensive answers or focuses on short or extractive responses.}
        \label{tab:data_comparison}
    \end{minipage}
\end{figure*}

\begin{figure*}[t]
    \centering
    \includegraphics[width=\textwidth]{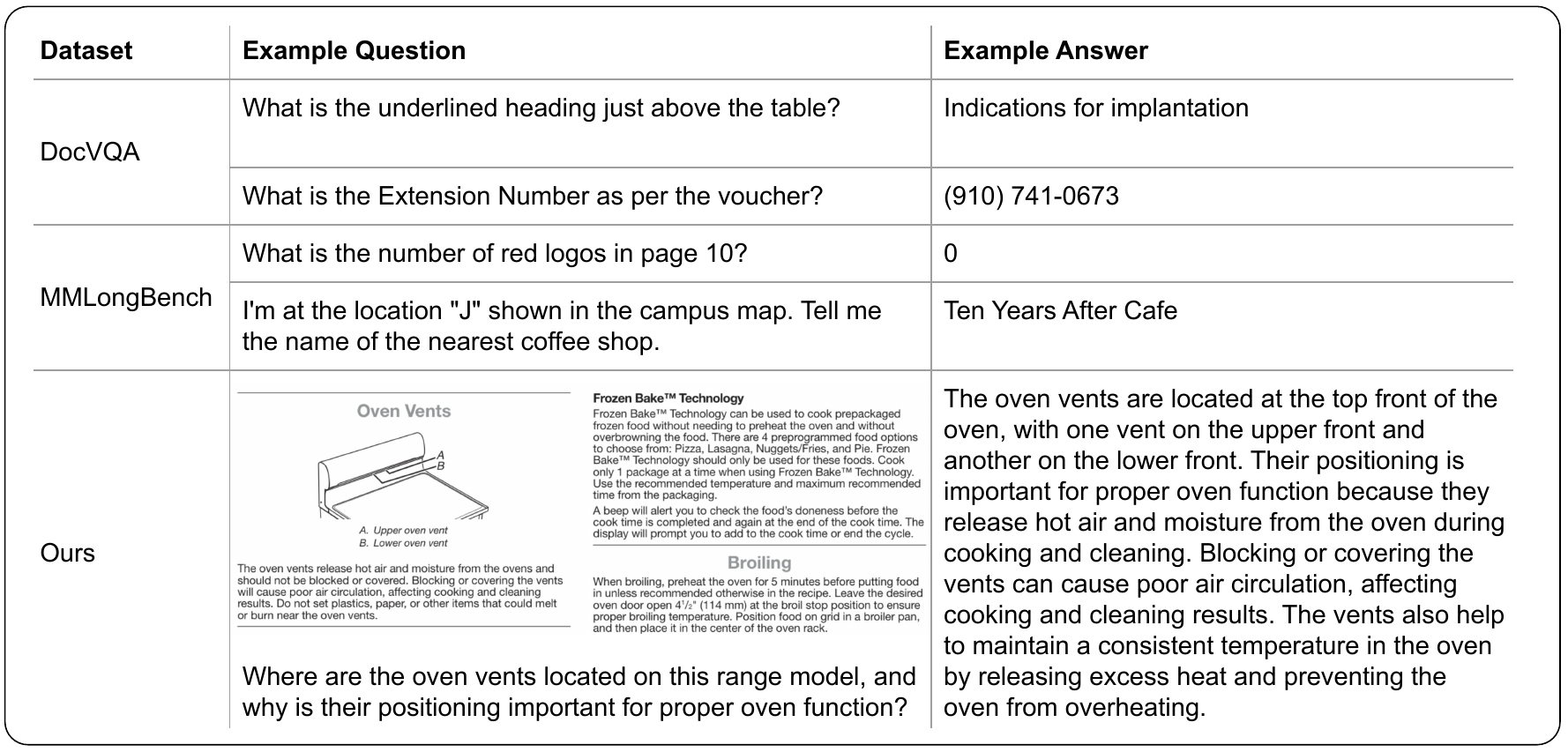}
    \caption{Example questions in different multimodal document question answering benchmarks. For illustration, we include content from the relevant page in the original document. 
    The example question from \datasetname{} is more complex than those from other benchmarks, as it requires an explanatory answer rather than an extraction of a short text span. Furthermore, it requires the model to understand the semantics of both image and text. 
    Please note that in our benchmark setting, the model is provided with all page contents from the document, and not only the relevant page. 
    }
    \label{fig:data_example}
\end{figure*}

To address these limitations, we introduce \datasetname{}, a comprehensive benchmark consisting of 
\datasize{}
samples specifically designed to evaluate the performance of large multimodal models on lengthy and diverse documents. 
Unlike previous datasets \citep{Mathew2020DocVQAAD, liu2024visualwebbench,  Ma2024MMLongBenchDocBL} 
that mainly contain short documents, \datasetname{} features recent documents \emph{spanning hundreds of pages, encompassing a wide range of topics and document structures} as shown in Figures \ref{fig:data_pie} 
and \ref{tab:data_comparison}. 
In addition, as shown in Figure \ref{fig:data_example}, our benchmark \emph{goes beyond simpler extractive questions, requiring models to provide open-ended solutions} that demonstrate in-depth understanding of the document content \citep{fan-etal-2019-eli5}.
\datasetname{} poses a question answering task where models have to analyze and reason over texts, figures, or tables in each multimodal long document.
We aim for this benchmark to serve as a valuable resource for researchers and practitioners, enabling more rigorous testing and development of multimodal document understanding systems.

{Another challenge we have to overcome is that due to the lengthy content of multimodal documents and the in-depth solutions, the evaluation of the open-ended question-answering task becomes tricky. 
To assess such open-ended solutions in a scalable and standardized manner, we design an automated evaluation framework that does not require reference answers or human annotation.
Inspired by previous works in model-based evaluation \citep{mt-judge-NEURIPS2023_91f18a12, zhao2024autoarenallmsautomating, liu-etal-2023-g-eval}, our evaluation framework leverages a detailed evaluation guide and multiple judge models to score the correctness of each generated solution.
}

{
With our proposed \datasetname{} and evaluation framework, we conducted
preliminary study on existing models and the results show that they struggle with figure and table-based questions compared to text-based questions, revealing their multimodal bias and weaknesses \citep{chen2024quantifyingmitigatingunimodalbiases}.
Furthermore, we observed that the models can be easily distracted by irrelevant content in the document pages \citep{irrelevant-context-10.5555/3618408.3619699}, even with the aid of retrieval-augmented generation.
}{
To enhance the robustness of multimodal models against potentially irrelevant retrieved content,
we propose a retrieval-aware tuning approach for multimodal document reading. 
This framework unifies supervised fine-tuning and retrieval augmented generation by including distracting content from other modalities and pages in each document. 
Thus, we adapt models to effectively incorporate the domain knowledge in multimodal documents while ignoring the content irrelevant to the given query. 
Compared to existing training methods, ours is the first to address retrieval-augmented multimodal training for rich document layouts.
To support this training framework and the enhancement of open-source models, we further construct a training corpus of \trainsize{} samples for question-answering over multimodal long documents.
}
Experiments show that our approach achieves a \performanceincrease{} relative improvement in the correctness of model responses, compared to the baseline model.

The key contributions of this work are threefold:
1) We introduce \datasetname{}, a multimodal benchmark that more accurately represents the challenges of real-world document understanding tasks. {Our automated evaluation framework enables scalable and standardized assessment of open-ended solutions.}
2) Our evaluation of leading models indicates that most models struggle with figure and table-based questions compared to text-based questions, revealing their multimodal bias.
3) We propose a retrieval-aware tuning framework that together with our large-scale training corpus, significantly improves the efficiency and effectiveness of multimodal document reading. 

Thus, we believe that this work contributes to the field of document understanding and paves the way for more capable and practical applications of large multimodal models in real-world scenarios. To accelerate the studies in our community, we will make the \datasetname{} benchmark, the training corpus for multimodal document reading, and our source code publicly available. 


\section{\datasetname{} Benchmark}

To evaluate the multimodal long document understanding ability of existing models, we present \datasetname{}, a challenging and diverse benchmark.
Notably, the benchmark focuses on open-ended questions that require in-depth solutions and analysis over very long documents with more than 200 pages on average.
For diversity, the questions cover the academic, financial, and product domains, with multiple topics in each domain.

\subsection{Data Collection}
\label{sec:data_collection}

To support our evaluation benchmark, we manually source high-quality multimodal documents from publicly accessible sources.
Concretely, we source research papers\footnote{\url{https://arxiv.org}}, company reports\footnote{\url{https://www.annualreports.com}} and product instruction manuals\footnote{\url{https://www.manualslib.com}} for the academic, financial, and product domains respectively.
Thus, the dataset covers a range of document formats and domains.
As research papers often require domain expertise, we constrain the academic domain to computer science topics.
To reduce the risk of data contamination or memorization when evaluating existing models \citep{dong-etal-2024-generalization-memorization}, we constrain the documents to be published in January 2024 or later.
As most existing models are unable to process raw PDF files, we conduct a simple data processing to extract the texts and relevant images from each document.
Specifically, we use the PyMuPDF\footnote{\url{https://pymupdf.readthedocs.io}} tool to automatically extract the text from each page.
To extract the figures and tables from each page, we leverage an existing object detection model \citep{doclaynet-10.1145/3534678.3539043}.
Thus, the processed documents consist of interleaved textual and visual content, where the visual contents are extracted images of figures and tables.

\begin{figure*}[t]
    \centering
    \includegraphics[width=\textwidth]{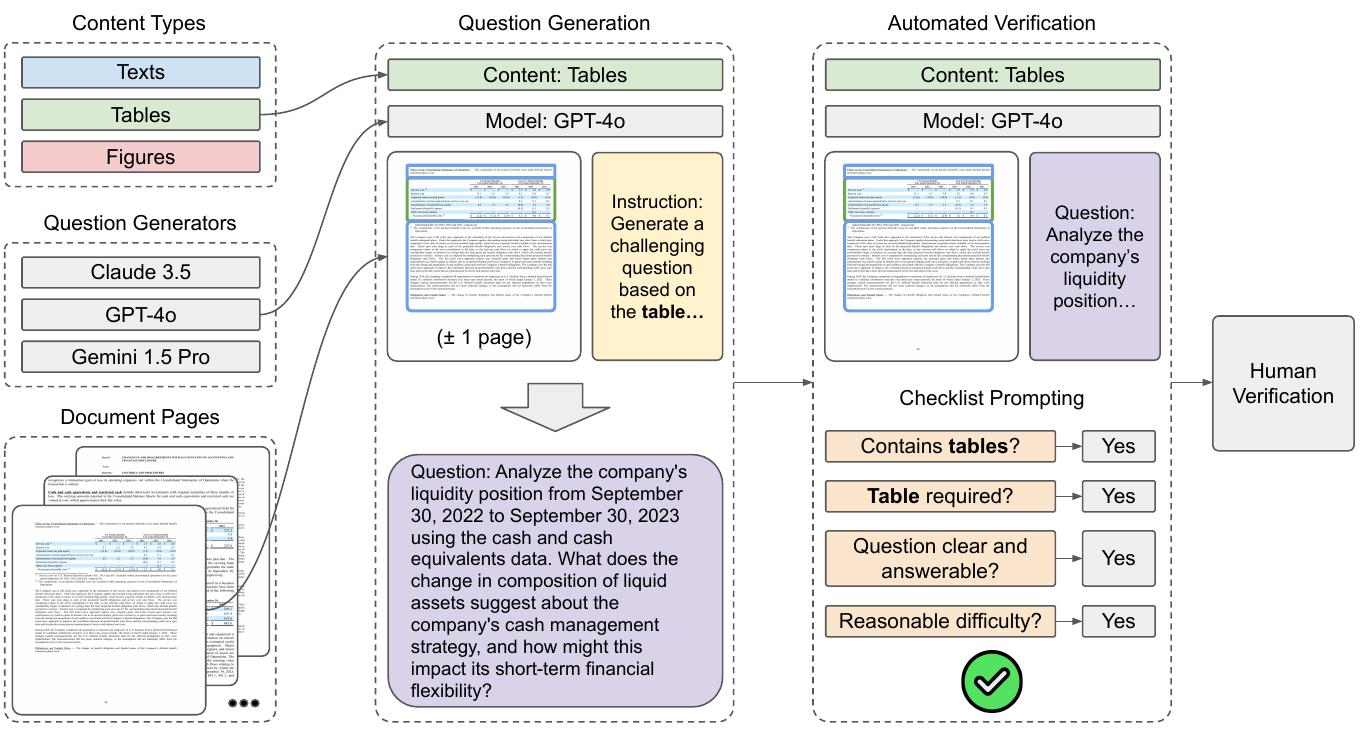}
    \caption{Overview of our data construction process with question verification stages. For brevity, we shorten the checklist prompts and include the full details in Appendix \ref{sec:annotation}. 
    }
    \label{fig:data_construction}
\end{figure*}

\subsection{Question Generation}

To construct diverse and challenging open-ended questions, we leverage a semi-automated pipeline.
Concretely, as shown in Figure \ref{fig:data_construction}, given a specified content category, we first randomly select a page from the document that contains the specific content category, such as texts, tables, or figures.
Consequently, we randomly select a question generator from a pool of leading multimodal models and instruct it to generate a challenging question based on the document page.
To ensure that the question generator has sufficient context, we also provide the previous page and subsequent page as additional inputs during the question generation process.

To improve the quality of the generated questions, we conduct an automated verification process as a preliminary filter for unsuitable questions. 
Concretely, the question generator is also instructed to reflect on the generated question and follow a multi-step checklist to validate the question.
For example, the checklist includes checking if the question is relevant to the document page, if the specified content category is required to answer the question, {and whether the question is answerable.}
The question is rejected if it does not satisfy any condition in the checklist.
Lastly, we employ a team of annotators to conduct final validation for each question. 
{We employ expert annotators who are Ph.D. students and above in computer science for the academic domain, and professional annotators for the finance and product domains.}
To be consistent, we provide a similar checklist and instruction as our automated verification stage, and the annotation details are included in Appendix \ref{sec:annotation}.
{We found that 80.1\% of the generated questions satisfied the automated verification. Of these questions that passed automated verification, 80.9\% also satisfied the human verification. Thus, we only retain \datasize{} questions that satisfied both the automated and human verification.} 

The statistics of our benchmark dataset are shown in Table \ref{tab:data_statistics}, where we ensure a balanced distribution of questions and documents from each domain and question category.
In this work, we focus on questions that require a single page of content to answer, 
and only retain answerable questions.
Please also note that while each question focuses on a specific category in a document page, the page may contain content from other categories as context.
For instance, a table-based question may also require comparisons to other tables or texts from the same page.

{Compared to the existing benchmarks in Figure \ref{tab:data_comparison}, \datasetname{} poses a greater challenge in two main aspects. Firstly, the significantly greater number of pages and tokens in each multimodal document poses extreme computational costs and opportunities to be distracted by irrelevant content \citep{irrelevant-context-10.5555/3618408.3619699}. While this challenge may be mitigated by retrieval-augmented generation \citep{chen-etal-2022-murag}, our preliminary study in Section \ref{sec:preliminary} shows that existing models are still hindered by their multimodal bias \citep{chen2024quantifyingmitigatingunimodalbiases}. In addition, our benchmark poses challenging open-ended questions as shown in Figure \ref{fig:data_example}, requiring models to produce in-depth analyses in their solutions.
Thus, we believe \datasetname{} is a more realistic and challenging benchmark compared with existing datasets focusing on short answers that can often be extracted directly from the source document.
}


\subsection{Automated Evaluation}
\label{sec:evaluation}

\begin{figure*}[t]
    \centering
    \includegraphics[width=0.95\textwidth]{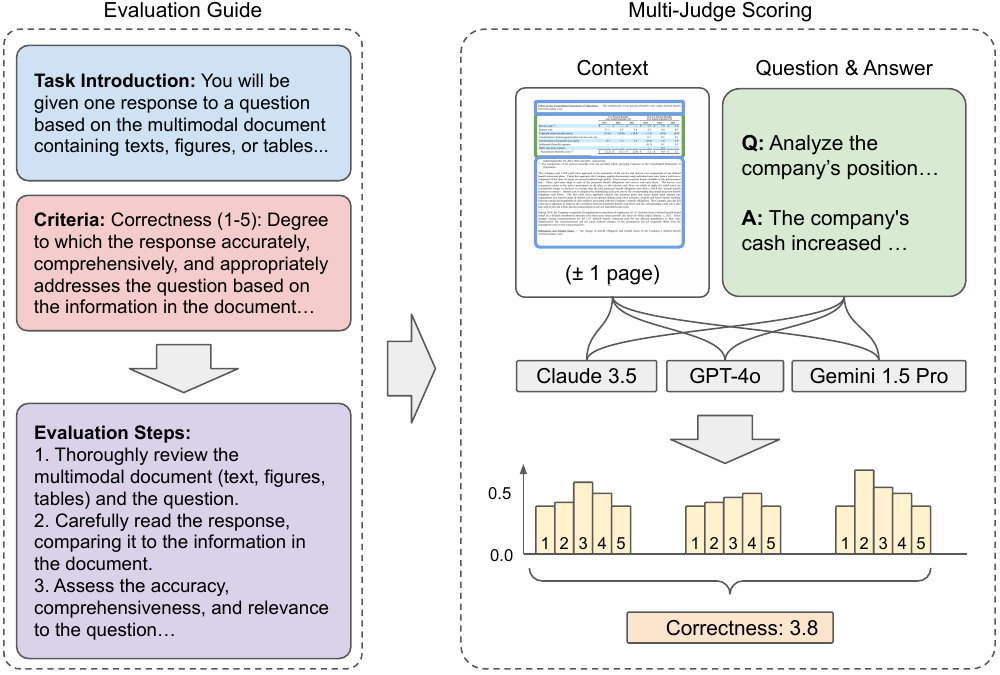}
    \caption{Our automated evaluation framework to assess the correctness of open-ended solutions for multimodal question answering. The full evaluation guide is included in Appendix \ref{sec:evaluation_guide}.
    }
    \label{fig:evaluation_framework}
\end{figure*}

Given the challenging nature of our multimodal long document benchmark, it is crucial to have a scalable and standardized evaluation method.
However, it is less feasible to conduct comprehensive human evaluation due to high labour costs and lack of reproducibility \citep{clark-etal-2021-thats-human}.
Thus, inspired by previous works in automatic evaluation 
\citep{mt-judge-NEURIPS2023_91f18a12, zhao2024autoarenallmsautomating, liu-etal-2023-g-eval}, we propose an evaluation framework based on a committee of multimodal judges.
Concretely, we leverage multiple leading multimodal models to score each answer to a question based on the criteria of correctness.
To provide a clear guideline for evaluation, we define the task introduction and criteria as shown in Figure \ref{fig:evaluation_framework}.
To provide more detailed evaluation instruction beyond the basic definitions above, we further construct detailed evaluation steps, based on the task and criteria.
Thus, the finalized evaluation guide for each judge model consists of the task introduction, criteria, and fixed evaluation steps. 

To provide a more reliable evaluation and reduce intra-model bias \citep{verga2024replacingjudgesjuriesevaluating}, we leverage multiple judges to evaluate each candidate answer. 
Specifically, each judge model $M_j$ is provided with the evaluation guide $g$, ground-truth evidence page as context $c$, question $q$, and candidate answer $\hat{a}$, and instructed to assign a correctness score from 1 to 5.
However, we observe some variance in the output scores, even with the same judge model and inputs.
Thus, we sample multiple scores from each judge model $M_j$ and aggregate the scores to obtain a fine-grained, continuous score that better reflects the quality of the candidate answer:
\begin{align}
    \text{Score} = \frac{1}{J \cdot K} \sum_{j=1}^J \sum_{k=1}^K s_{j,k} \sim M_{j}(g, c, q, \hat{a})
\end{align}
where $J=3$ is the number of judge models and $K=5$ is the number of sampled scores per judge model.
While there may be some degree of subjectiveness in our framework, our analysis in later sections shows that it largely agrees with human preferences with minimal bias \citep{mt-judge-NEURIPS2023_91f18a12}.
Thus, we believe this automated evaluation framework is reliable and more scalable.

\begin{table}[t]
\centering
\begin{tabular}{lcccc}
\toprule
& Academic & Product & Financial & All \\
& Paper & Manuals & Report & \\
\midrule
Documents & 60 & 60 & 60 & 180 \\


Questions & 311 & 279 & 261 & 851 \\
\quad Text-based questions & 95 & 95 & 81 & 271 \\
\quad Figure-based questions & 114 & 93 & 76 & 283 \\
\quad Table-based questions & 102 & 91 & 104 & 297 \\

Average pages per document & 201.2 & 277.8 & 153.4 & 210.8 \\
Average text tokens per document & 114,129.8 & 109,745.0 & 139,089.3 & 120,988.0 \\
Average figure images per document & 90.8 & 368.3 & 24.1 & 161.13 \\
Average table images per document & 34.9 & 96.6 & 83.8 & 71.8 \\
\bottomrule
\end{tabular}
\caption{Benchmark dataset statistics with respect to each domain. 
}
\label{tab:data_statistics}
\end{table}

\subsection{Preliminary Study}
\label{sec:preliminary}

\begin{table}[ht]
\centering
\begin{tabular}{lcccc}
\toprule
& Text & Figure & Table & All \\
\midrule
Gemini-1.5-pro-002 \\
\quad w/ top $k=1$ pages & 4.38 & 3.73 & 4.16 & 4.11 \\
\quad w/ top $k=5$ pages & 4.60 & 4.31 & 4.54 & 4.49 \\
\quad w/ top $k=10$ pages & 4.61 & 4.29 & 4.62 & 4.51 \\
\quad w/ top $k=20$ pages & 4.63 & 4.33 & 4.38 & 4.46 \\
\midrule
Qwen2-VL-7B-Instruct \\
\quad w/ top $k=1$ pages & 4.05 & 3.25 & 3.36 & 3.57 \\
\quad w/ top $k=5$ pages & 4.17 & 3.67 & 3.46 & 3.78 \\
\quad w/ top $k=10$ pages & 4.08 & 3.62 & 3.19 & 3.65 \\
\quad w/ top $k=20$ pages & OOM & OOM & OOM & OOM \\
\bottomrule
\end{tabular}
\caption{Preliminary study on \datasetname{} for open-source and close-source models. We report the correctness score out of 5 for text-based, figure-based, table-based, and all questions respectively.}
\label{tab:preliminary}
\end{table}

To investigate the limitations of existing models, we conduct a preliminary study on a subset of 100 random samples from our \datasetname{} benchmark.
Concretely, we select Gemini \citep{reid2024gemini15} and Qwen2-VL \citep{wang2024qwen2vlenhancingvisionlanguagemodels} to represent highly capable models for the close-source and open-source settings respectively.
While large multimodal models have shown impressive capabilities and support longer input contexts,
they often struggle with understanding very long documents and may incur great computational costs \citep{dingjie2024milebench}.
Thus, we focus our study on the retrieval-augmented generation paradigm \citep{NEURIPS2020_6b493230_rag}, which leverages a retriever to select only the most relevant content, and the retrieved content is used to augment the generator model inputs for question answering.
Concretely, we use ColPali \citep{Faysse2024ColPaliED} as a state-of-the-art multimodal retriever and leverage the top $k$ pages of multimodal content as context.
We include more details on the implementation and evaluation of retriever performance in Appendix \ref{sec:retrieval}.

Notably, as shown in Table \ref{tab:preliminary}, we observe significantly lower performance for figure-based and table-based questions, as compared to text-based questions.
We believe that this discrepancy suggests that current models are weaker in processing image-based contents in multimodal documents, or may be biased towards the textual content, even when they are trained on interleaved multimodal data \citep{chen2024quantifyingmitigatingunimodalbiases}.
Furthermore, we find that increasing the amount of retrieved content may not improve overall performance, and may even lead to worse performance or out-of-memory (OOM) issues.
This indicates that the multimodal generator models may be easily distracted by irrelevant content in longer contexts \citep{irrelevant-context-10.5555/3618408.3619699}.
Thus, we believe it is crucial to address these challenges of existing models for processing multimodal long documents.

{
Additionally, to verify the reliability of our automated evaluation, we conduct manual human scoring based on the same evaluation guide.
For the samples in this preliminary study, we observed a Pearson correlation of 88.9\% with $p < 0.001$ between the final aggregated score from the judge models, and the human annotator.
Thus, we believe that our evaluation framework can achieve a very high agreement with human preferences despite the open-ended and in-depth nature of the answers.
}

\section{Retrieval-Aware Multimodal Tuning}
\label{sec:method}

\begin{figure*}[t]
    \centering
    \includegraphics[width=\textwidth]{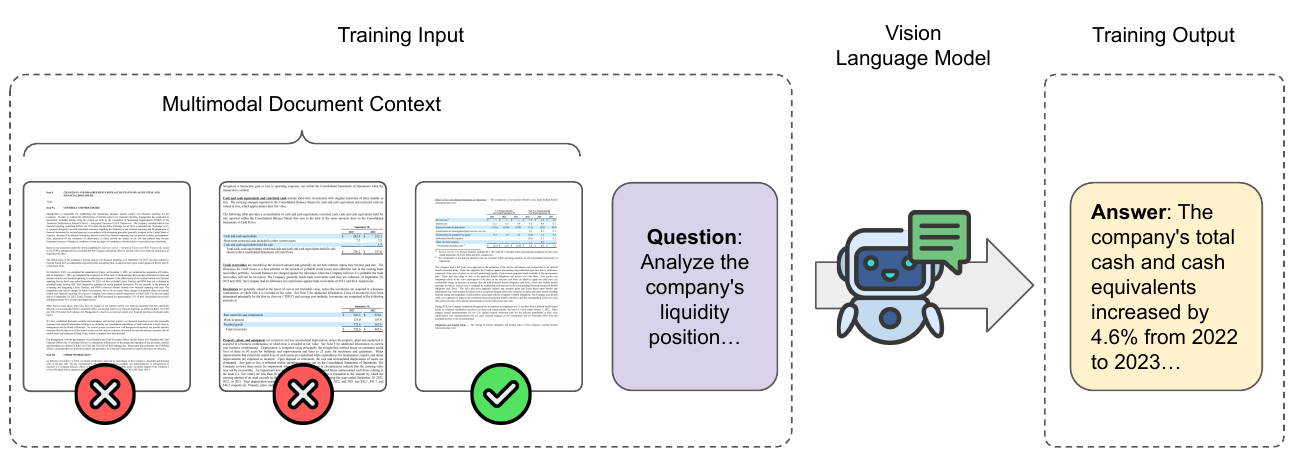}
    \caption{Our retrieval-aware multimodal tuning framework to enhance the ability of models to identify and utilize pertinent content in multimodal documents. At training time, the model is provided with more relevant pages retrieved from the document, which may contain both the gold evidence page and multiple `distractor' pages. 
    }
    \label{fig:training}
\end{figure*}

While current multimodal models are often trained on interleaved inputs with multiple pieces of texts and images \citep{liu2023visualllava}, they may not be well-optimized for multimodal documents. Specifically, multimodal documents are more challenging to understand as they contain diverse multimodal content including unstructured texts, and images representing structured tables and figures.
Thus, models require a more fine-grained understanding and perception of the multimodal document content.
Furthermore, the retrieval setting requires models to ground their outputs in the additional context by identifying and utilizing 
only the pertinent content. 
However, previous works \citep{irrelevant-context-10.5555/3618408.3619699} and our preliminary study have shown that they may still be easily misled by some irrelevant information in the retrieved content.
To our knowledge, there is no open-source model that can address these challenges in multimodal long documents.

To this end, we propose a simple and effective retrieval-aware multimodal document tuning approach. 
Inspired by previous works in retrieval augmentation \citep{chen-etal-2022-murag, zhang2024raft}, we include both retrieval context from the ground-truth evidence page as well as potentially irrelevant pages during training.
Our approach as shown in Figure \ref{fig:training} presents a training paradigm that is more realistic and similar to the challenges faced during test-time retrieval of multimodal content.
Thus, the model learns to handle potentially noisy retrieval contexts, while improving its text understanding and visual perception capabilities to utilize the most relevant document content.
To construct the training data, we leverage the same process as shown in Figure \ref{fig:data_construction} to construct a training corpus of \trainsize{} samples across 300 documents, and leverage the respective question generator models to also produce a high-quality answer based on each ground-truth evidence page.
We omit the human verification stage for scalability and cost-efficiency, as majority of the automatically verified samples also satisfied human verification. 
To assess the quality of the generated training solutions, we evaluated using our automated scoring framework on a random subset of 100 samples.
We observed an average correctness score of 4.82, indicating very high quality of answers in the training data.
To avoid data leakage, we ensure that the documents used to construct the training corpus do not overlap with the evaluation set. 
For example, we ensure that the training documents are from different companies and products, and are published in an earlier time period. 


\section{Experiments}
\label{sec:experiments}

\subsection{Task Setting}

To ensure a practical task setting, we focus on the retrieval-based paradigm, which avoids the exorbitant cost to process the full document.
Based on our preliminary study in Section \ref{sec:preliminary}, we use the top $k=5$ pages ranked by the retriever as a reasonable amount of context for each question.
Thus, each model is provided with the retrieved context and question as input, and required to provide an open-ended solution as output.
As discussed in Section \ref{sec:evaluation}, we leverage an automated framework with multiple judge models to score the correctness of each output solution, on a scale of 1 to 5.

\subsection{Models}

To provide a more comprehensive investigation of current models, we use both open-source and close-source models in this work.
Concretely, we select GPT-4o (gpt-4o-2024-05-13)\footnote{\url{https://openai.com/index/gpt-4o-system-card/}}, Claude 3.5 Sonnet (claude-3-5-sonnet-20240620)\footnote{\url{https://www.anthropic.com/news/claude-3-5-sonnet}} and Gemini 1.5 Pro (gemini-1.5-pro-002) \citep{reid2024gemini15} due to their leading performance on multimodal benchmarks \citep{yue2023mmmu}.
Regarding open-source models, we specifically select models which support interleaved multimodal inputs with multiple images, and fine-grained visual perception of document content.
Thus, we mainly focus on LLaVA-OneVision-7B \citep{li2024llavaonevisioneasyvisualtask} and Qwen2-VL-7B-Instruct \citep{wang2024qwen2vlenhancingvisionlanguagemodels}.
We plan to expand our investigation to other capable open-source models as they are released.

\subsection{Hyperparameters}

For all models, we use greedy decoding with temperature $T=0$ to reduce variance. 
In our training framework, we set the number of training epochs to be 1, batch size as 16, and learning rate as 1e-4.
To reduce the training cost due to limited computational resources, we leverage LoRA \citep{hu2022lora} training with rank as 64 and alpha as 32.
Due to training instabilities with other open-source models, we mainly focus the training experiments on the Qwen2-VL-7B-Instruct model, which demonstrates leading performance compared to similar-sized models.

\begin{table}[t]
\centering
\begin{tabular}{llccccccc}
\toprule
& & \multicolumn{3}{c}{Domain} & \multicolumn{3}{c}{Question Category} \\
\cmidrule(lr){3-5} \cmidrule(lr){6-8}
Model & Size & Academic & Product & Finance & Text & Figure & Table & All \\
\midrule
\multicolumn{9}{c}{\textit{Proprietary Models}} \\
\midrule
GPT-4o & - & 4.56 & 4.38 & 4.51 & 4.55 & 4.38 & 4.53 & 4.49 \\
Claude 3.5 Sonnet & - & 4.59 & 4.43 & 4.51 & 4.57 & 4.42 & 4.54 & 4.51 \\
Gemini 1.5 Pro & - & 4.66 & 4.43 & 4.43 & 4.59 & 4.43 & 4.52 & 4.51 \\
\midrule
\multicolumn{9}{c}{\textit{Open-Source Models}} \\
\midrule
LLaVA OneVision & 7B & 3.71 & 3.74 & 3.39 & 4.03 & 3.57 & 3.30 & 3.62 \\
Qwen2-VL & 7B & 4.03 & 3.88 & 3.56 & 4.08 & 3.83 & 3.62 & 3.84 \\
\quad w/ Retrieval Tuning & 7B &  \textbf{4.17} & \textbf{4.01} & \textbf{3.86} & \textbf{4.31}     & \textbf{4.00}    & \textbf{3.77} & \textbf{4.02} \\
\bottomrule
\end{tabular}
\caption{Evaluation of model performance for proprietary and open-source multimodal models. We report the correctness on our benchmark across different document domains and question categories. We bold the highest scores obtained by open-source models. }
\label{tab:main_results}
\end{table}

\section{Results}

\subsection{Main Results}
To assess the effectiveness of our approach and the holistic performance of existing models, we report the main evaluation results in Table \ref{tab:main_results}.
First, we find that our retrieval-aware multimodal tuning significantly and consistently enhances the performance of Qwen2-VL, representing a relative improvement of  \performanceincrease{} in answer correctness. Thus, we view the proposed training approach as a promising strategy to enhance multimodal long document understanding ability, and reduce the gap between open-source and proprietary models.
Second, we observe that open-source models have worse performance in answering table-related questions compared to other question categories. This discrepancy highlights the need for more efforts to enhance the table understanding capability of open-source multimodal models. 
We include further qualitative analysis of the model predictions in the Appendix \ref{sec:more_examples}.

\begin{table}[t]
\centering
\begin{tabular}{lccc}
\toprule
& \multicolumn{3}{c}{Question Category} \\
\cmidrule(lr){2-4}
Model & Text & Figure & Table \\
\midrule
Qwen2-VL & 4.08 & 3.83 & 3.62 \\
\quad w/o Image Inputs & 4.22 & 3.37 & 3.38 \\
\quad w/ Render Page as Inputs & 3.99 & 3.70 & 3.39 \\
\bottomrule
\end{tabular}
\caption{Analysis on alternative settings for our benchmark, including removing images from model inputs, and using only the render image of each page as document context, without text extraction.}
\label{tab:alternative_settings}
\end{table}

\subsection{Effect of Alternative Settings}

While we mainly focus on the multimodal setting with extracted texts and images, we believe it is also important to explore other settings in practice.
{Our main data setting as discussed in Section \ref{sec:data_collection} first extracts the texts, figures, and tables separately, with the figures and tables represented as individual images.}
As shown in Table \ref{tab:alternative_settings}, we find a significant decrease in performance for figure-based and table-based questions when the image inputs are removed, although the model may still be able to answer the questions to an extent as the extracted text may contain partial information about the tables and figures.
We believe that this underscores the importance of leveraging multimodal content in documents, even though many documents may contain a majority of the content as texts.
The performance increases slightly for text-based questions when image inputs are removed, suggesting that the images may mislead the model in rare cases.
However, we believe this is acceptable as questions may cover a wide variety of multimodal content in practice.
On the other hand, we observe that it may be less optimal to use only rendered images of document pages as inputs, instead of separately extracting the texts, tables, and figures as in our main setting.
{While the rendered page image does contain the original information and layout of the document, including texts, tables, and figures, the model may be less capable of distinguishing the content between texts and tables.}

\section{Related Work}

\subsection{Large Multimodal Models}

In recent years, large multimodal models have demonstrated their capability to process and comprehend data across various formats. 
Close-source models such as GPT-4o \citep{gpt4o} can reason across audio, vision and text. 
Claude 3.5 Sonnet \citep{claude35} shows marked improvement on tasks that require visual reasoning like interpreting charts and graphs. 
On the other hand, Gemini 1.5 Pro \citep{reid2024gemini15} is capable of reasoning over multiple long documents and hours of video and audio.
Open-source models such as Llava \citep{liu2023visualinstructiontuning}, 
Idefics \citep{laurencon2023obelics},
Otter \citep{li2023ottermultimodalmodelincontext},
InternVL \citep{chen2024far},
CogVLM \citep{wang2023cogvlm}, have also shown the potential over various types of multi-modal content including document images~\citep{Mathew2020DocVQAAD}, slides~\citep{Tanaka_Nishida_Nishida_Hasegawa_Saito_Saito_2023_slidevqa}, and charts~\citep{Huang-etal-2024-chart-survey}. 
However, the benchmark performance of open-source models tends to lag behind that of close-source models \citep{yue2023mmmu}, prompting an urgent need to bridge the gap.
In this work, we introduce a retrieval-aware multimodal tuning framework which can significantly improve the multimodal long document understanding ability of models.


\subsection{Document Understanding Datasets}

Given the practical and business applications of document understanding, researchers have devoted significant effort to this area by introducing new datasets and methods.
SearchQA \citep{dunn2017searchqa}, NarrativeQA \citep{kovcisky2018narrativeqa}, QuALITY \citep{zhu2020question} are reading comprehension datasets over purely textual data with an average length ranging from 1850 to 60k tokens.
FinQA \citep{Chen2021FinQAAD}, DocFinQA \citep{Reddy2024DocFinQAAL} are introduced in the financial domain.
MarkQA \citep{Huang2023MarkQAAL} tackes QA over knowledge bases with numerical reasoning.
DocVQA\citep{Mathew2020DocVQAAD} presents a visual question answering dataset on document images.
VisualWebBench \citep{liu2024visualwebbench} is a multimodal benchmark over single-page documents focusing on various QA-style tasks.
MMLongBench
\citep{Ma2024MMLongBenchDocBL} is a multimodal document understanding dataset with an average of 47.5 pages and 21k textual tokens.
Methods such as PDFTriage \citep{saad2023pdftriage} enables models to retrieve the context from long and structured documents. 
TAT-LLM \citep{zhu2024tat} addresses QA over a hybrid of tabular and textual data.
ChartQA~\citep{masry-etal-2022-chartqa} is a benchmark of extractive questions-answering task over a chart image, while the Chocolate dataset~\citep{charts-huang-etal-2024-lvlms} annotates the types of factual errors in machine-generated chart captions. 
Compared to the datasets above, our \datasetname{} benchmark contains more lengthy documents with hundreds of pages and focusing on open-ended questions which require in-depth solutions.
We further propose an automated and reliable evaluation framework to assess the correctness of model answers, which demonstrates very high agreement with human preferences.





\subsection{Retrieval-Augmented Generation}

While recent multimodal models have shown impressive capability in many tasks, applying them directly to long document understanding tasks may face several challenges to the diverse multimodal content.
Additionally, processing entire documents with large models is often impractical, as the text alone may contain millions of tokens, leading to substantial computational costs.
Therefore, researchers have designed various retrieval augmented generation \citep{NEURIPS2020_6b493230_rag, chen-etal-2022-murag} methods to address the issues.
In this work, we have investigated multiple retrieval methods optimized for document page retrieval, including JINA-CLIP \citep{xiao2024jina}
BM25 \citep{Robertson2009ThePRbm25}, BGE-M3 \citep{chen2024bgem3embeddingmultilingualmultifunctionality} and ColPali \citep{Faysse2024ColPaliED}.
However, despite using retrieval, existing multimodal models are constrained by their multimodal biases \citep{chen2024quantifyingmitigatingunimodalbiases} and susceptibility to irrelevant content that is inherent in the retrieved context \citep{irrelevant-context-10.5555/3618408.3619699}. 
Thus, we further proposed a retrieval-aware tuning framework to enhance the performance of models when leveraging retrieval for multimodal long documents.




\section{Conclusion}

In this work, we introduce \datasetname{}, a benchmark dataset consisting of \datasize{} samples and an automated framework to evaluate the performance of large multimodal models on document question answering tasks. This benchmark is specifically designed for long and diverse document formats containing text, figures, and tables, aligning with the demands of real-world applications. Unlike existing benchmarks, \datasetname{} features more recent and lengthy documents, often hundreds of pages long, and requires open-ended solutions rather than just extractive answers.

We also propose a retrieval-aware tuning approach designed for the efficient and effective processing of multimodal long documents. To our knowledge, this is the first training framework and model to directly address the retrieval setting for such documents.
Experimental results demonstrate that our method achieves a relative improvement of \performanceincrease{} in the correctness of model responses compared to baseline open-source models. This improvement showcases the effectiveness of our approach in handling lengthy and complex multimodal documents, potentially aiding humans in various business and practical applications that require understanding and answering questions over such documents.


\section*{Acknowledgment}

This work was substantially supported by DAMO Academy through DAMO Academy Research Intern Program.

\section*{Ethics Statement}

We will release the benchmark and training dataset publicly to facilitate further research in this area.
To observe copyright rules, we do not release the documents directly, but instead the links to download each document.
All annotators in this work were volunteers.
While we focus on how models may answer questions based on multimodal documents, it is still possible for them to hallucinate information that is false or not verifiable.

\section*{Reproducibility Statement}

In this work, we have included the details of our training framework and hyperparameters in Section \ref{sec:method} and \ref{sec:experiments}.
As discussed above, our benchmark dataset and questions will be released under a public licence.
For reproducibility, our code will be found at \url{https://anonymous.4open.science/r/private-multimodal-documents-B2CF/}.

\bibliography{custom}

\begin{thebibliography}{48}
\providecommand{\natexlab}[1]{#1}
\providecommand{\url}[1]{\texttt{#1}}
\expandafter\ifx\csname urlstyle\endcsname\relax
  \providecommand{\doi}[1]{doi: #1}\else
  \providecommand{\doi}{doi: \begingroup \urlstyle{rm}\Url}\fi

\bibitem[AI(2024)]{gpt4o}
Open AI.
\newblock Hello gpt-4o, 2024.
\newblock URL \url{https://openai.com/index/hello-gpt-4o/}.

\bibitem[Anthropic(2024)]{claude35}
Anthropic.
\newblock Claude 3.5 sonnet, 2024.
\newblock URL \url{https://www.anthropic.com/news/claude-3-5-sonnet}.

\bibitem[Chen et~al.(2024{\natexlab{a}})Chen, Xiao, Zhang, Luo, Lian, and Liu]{chen2024bgem3embeddingmultilingualmultifunctionality}
Jianlv Chen, Shitao Xiao, Peitian Zhang, Kun Luo, Defu Lian, and Zheng Liu.
\newblock Bge m3-embedding: Multi-lingual, multi-functionality, multi-granularity text embeddings through self-knowledge distillation, 2024{\natexlab{a}}.
\newblock URL \url{https://arxiv.org/abs/2402.03216}.

\bibitem[Chen et~al.(2024{\natexlab{b}})Chen, Cao, Zhang, and Lu]{chen2024quantifyingmitigatingunimodalbiases}
Meiqi Chen, Yixin Cao, Yan Zhang, and Chaochao Lu.
\newblock Quantifying and mitigating unimodal biases in multimodal large language models: A causal perspective, 2024{\natexlab{b}}.
\newblock URL \url{https://arxiv.org/abs/2403.18346}.

\bibitem[Chen et~al.(2022)Chen, Hu, Chen, Verga, and Cohen]{chen-etal-2022-murag}
Wenhu Chen, Hexiang Hu, Xi~Chen, Pat Verga, and William Cohen.
\newblock {M}u{RAG}: Multimodal retrieval-augmented generator for open question answering over images and text.
\newblock In Yoav Goldberg, Zornitsa Kozareva, and Yue Zhang (eds.), \emph{Proceedings of the 2022 Conference on Empirical Methods in Natural Language Processing}, pp.\  5558--5570, Abu Dhabi, United Arab Emirates, December 2022. Association for Computational Linguistics.
\newblock \doi{10.18653/v1/2022.emnlp-main.375}.
\newblock URL \url{https://aclanthology.org/2022.emnlp-main.375}.

\bibitem[Chen et~al.(2024{\natexlab{c}})Chen, Wang, Tian, Ye, Gao, Cui, Tong, Hu, Luo, Ma, et~al.]{chen2024far}
Zhe Chen, Weiyun Wang, Hao Tian, Shenglong Ye, Zhangwei Gao, Erfei Cui, Wenwen Tong, Kongzhi Hu, Jiapeng Luo, Zheng Ma, et~al.
\newblock How far are we to gpt-4v? closing the gap to commercial multimodal models with open-source suites.
\newblock \emph{arXiv preprint arXiv:2404.16821}, 2024{\natexlab{c}}.
\newblock URL \url{https://arxiv.org/abs/2404.16821}.

\bibitem[Chen et~al.(2021)Chen, Chen, Smiley, Shah, Borova, Langdon, Moussa, Beane, Huang, Routledge, and Wang]{Chen2021FinQAAD}
Zhiyu Chen, Wenhu Chen, Charese Smiley, Sameena Shah, Iana Borova, Dylan Langdon, Reema~N Moussa, Matthew~I. Beane, Ting-Hao~'Kenneth' Huang, Bryan~R. Routledge, and William~Yang Wang.
\newblock Finqa: A dataset of numerical reasoning over financial data.
\newblock In \emph{Proceedings of EMNLP}, 2021.
\newblock URL \url{https://api.semanticscholar.org/CorpusID:235399966}.

\bibitem[Clark et~al.(2021)Clark, August, Serrano, Haduong, Gururangan, and Smith]{clark-etal-2021-thats-human}
Elizabeth Clark, Tal August, Sofia Serrano, Nikita Haduong, Suchin Gururangan, and Noah~A. Smith.
\newblock All that{'}s {`}human{'} is not gold: Evaluating human evaluation of generated text.
\newblock In Chengqing Zong, Fei Xia, Wenjie Li, and Roberto Navigli (eds.), \emph{Proceedings of the 59th Annual Meeting of the Association for Computational Linguistics and the 11th International Joint Conference on Natural Language Processing (Volume 1: Long Papers)}, pp.\  7282--7296, Online, August 2021. Association for Computational Linguistics.
\newblock \doi{10.18653/v1/2021.acl-long.565}.
\newblock URL \url{https://aclanthology.org/2021.acl-long.565}.

\bibitem[Dingjie et~al.(2024)Dingjie, Chen, Chen, Yu, Wan, and Wang]{dingjie2024milebench}
Song Dingjie, Shunian Chen, Guiming~Hardy Chen, Fei Yu, Xiang Wan, and Benyou Wang.
\newblock Milebench: Benchmarking {MLLM}s in long context.
\newblock In \emph{First Conference on Language Modeling}, 2024.
\newblock URL \url{https://openreview.net/forum?id=Uhwze2LEwq}.

\bibitem[Dong et~al.(2024)Dong, Jiang, Liu, Jin, Gu, Yang, and Li]{dong-etal-2024-generalization-memorization}
Yihong Dong, Xue Jiang, Huanyu Liu, Zhi Jin, Bin Gu, Mengfei Yang, and Ge~Li.
\newblock Generalization or memorization: Data contamination and trustworthy evaluation for large language models.
\newblock In Lun-Wei Ku, Andre Martins, and Vivek Srikumar (eds.), \emph{Findings of the Association for Computational Linguistics ACL 2024}, pp.\  12039--12050, Bangkok, Thailand and virtual meeting, August 2024. Association for Computational Linguistics.
\newblock \doi{10.18653/v1/2024.findings-acl.716}.
\newblock URL \url{https://aclanthology.org/2024.findings-acl.716}.

\bibitem[Dunn et~al.(2017)Dunn, Sagun, Higgins, Guney, Cirik, and Cho]{dunn2017searchqa}
Matthew Dunn, Levent Sagun, Mike Higgins, V~Ugur Guney, Volkan Cirik, and Kyunghyun Cho.
\newblock Searchqa: A new q\&a dataset augmented with context from a search engine.
\newblock \emph{arXiv preprint arXiv:1704.05179}, 2017.
\newblock URL \url{https://arxiv.org/pdf/1704.05179}.

\bibitem[Fan et~al.(2019)Fan, Jernite, Perez, Grangier, Weston, and Auli]{fan-etal-2019-eli5}
Angela Fan, Yacine Jernite, Ethan Perez, David Grangier, Jason Weston, and Michael Auli.
\newblock {ELI}5: Long form question answering.
\newblock In Anna Korhonen, David Traum, and Llu{\'\i}s M{\`a}rquez (eds.), \emph{Proceedings of the 57th Annual Meeting of the Association for Computational Linguistics}, pp.\  3558--3567, Florence, Italy, July 2019. Association for Computational Linguistics.
\newblock \doi{10.18653/v1/P19-1346}.
\newblock URL \url{https://aclanthology.org/P19-1346}.

\bibitem[Faysse et~al.(2024)Faysse, Sibille, Wu, Omrani, Viaud, Hudelot, and Colombo]{Faysse2024ColPaliED}
Manuel Faysse, Hugues Sibille, Tony Wu, Bilel Omrani, Gautier Viaud, C'eline Hudelot, and Pierre Colombo.
\newblock Colpali: Efficient document retrieval with vision language models.
\newblock \emph{ArXiv}, abs/2407.01449, 2024.
\newblock URL \url{https://api.semanticscholar.org/CorpusID:270870444}.

\bibitem[Fujitake(2024)]{fujitake-2024-layoutllm}
Masato Fujitake.
\newblock {L}ayout{LLM}: Large language model instruction tuning for visually rich document understanding.
\newblock In Nicoletta Calzolari, Min-Yen Kan, Veronique Hoste, Alessandro Lenci, Sakriani Sakti, and Nianwen Xue (eds.), \emph{Proceedings of the 2024 Joint International Conference on Computational Linguistics, Language Resources and Evaluation (LREC-COLING 2024)}, pp.\  10219--10224, Torino, Italia, May 2024. ELRA and ICCL.
\newblock URL \url{https://aclanthology.org/2024.lrec-main.892}.

\bibitem[Google(2024)]{reid2024gemini15}
Gemini~Team Google.
\newblock Gemini 1.5: Unlocking multimodal understanding across millions of tokens of context.
\newblock \emph{arXiv preprint arXiv:2403.05530}, 2024.

\bibitem[Hu et~al.(2022)Hu, yelong shen, Wallis, Allen-Zhu, Li, Wang, Wang, and Chen]{hu2022lora}
Edward~J Hu, yelong shen, Phillip Wallis, Zeyuan Allen-Zhu, Yuanzhi Li, Shean Wang, Lu~Wang, and Weizhu Chen.
\newblock Lo{RA}: Low-rank adaptation of large language models.
\newblock In \emph{International Conference on Learning Representations}, 2022.
\newblock URL \url{https://openreview.net/forum?id=nZeVKeeFYf9}.

\bibitem[Huang et~al.(2024{\natexlab{a}})Huang, Chan, Fung, Qiu, Zhou, Joty, Chang, and Ji]{Huang-etal-2024-chart-survey}
Kung{-}Hsiang Huang, Hou~Pong Chan, Yi~Ren Fung, Haoyi Qiu, Mingyang Zhou, Shafiq Joty, Shih{-}Fu Chang, and Heng Ji.
\newblock From pixels to insights: {A} survey on automatic chart understanding in the era of large foundation models.
\newblock \emph{CoRR}, abs/2403.12027, 2024{\natexlab{a}}.
\newblock \doi{10.48550/ARXIV.2403.12027}.
\newblock URL \url{https://doi.org/10.48550/arXiv.2403.12027}.

\bibitem[Huang et~al.(2024{\natexlab{b}})Huang, Zhou, Chan, Fung, Wang, Zhang, Chang, and Ji]{charts-huang-etal-2024-lvlms}
Kung-Hsiang Huang, Mingyang Zhou, Hou~Pong Chan, Yi~Fung, Zhenhailong Wang, Lingyu Zhang, Shih-Fu Chang, and Heng Ji.
\newblock Do {LVLM}s understand charts? analyzing and correcting factual errors in chart captioning.
\newblock In Lun-Wei Ku, Andre Martins, and Vivek Srikumar (eds.), \emph{Findings of the Association for Computational Linguistics ACL 2024}, pp.\  730--749, Bangkok, Thailand and virtual meeting, August 2024{\natexlab{b}}. Association for Computational Linguistics.
\newblock \doi{10.18653/v1/2024.findings-acl.41}.
\newblock URL \url{https://aclanthology.org/2024.findings-acl.41}.

\bibitem[Huang et~al.(2023)Huang, Cheng, Bao, Huang, and Qu]{Huang2023MarkQAAL}
Xiang Huang, Sitao Cheng, Yuheng Bao, Shanshan Huang, and Yuzhong Qu.
\newblock Markqa: A large scale kbqa dataset with numerical reasoning.
\newblock In \emph{Proceedings of EMNLP}, 2023.
\newblock URL \url{https://api.semanticscholar.org/CorpusID:264439563}.

\bibitem[Ko{\v{c}}isk{\`y} et~al.(2018)Ko{\v{c}}isk{\`y}, Schwarz, Blunsom, Dyer, Hermann, Melis, and Grefenstette]{kovcisky2018narrativeqa}
Tom{\'a}{\v{s}} Ko{\v{c}}isk{\`y}, Jonathan Schwarz, Phil Blunsom, Chris Dyer, Karl~Moritz Hermann, G{\'a}bor Melis, and Edward Grefenstette.
\newblock The narrativeqa reading comprehension challenge.
\newblock \emph{TACL}, 6:\penalty0 317--328, 2018.

\bibitem[Laurençon et~al.(2023)Laurençon, Saulnier, Tronchon, Bekman, Singh, Lozhkov, Wang, Karamcheti, Rush, Kiela, Cord, and Sanh]{laurencon2023obelics}
Hugo Laurençon, Lucile Saulnier, Léo Tronchon, Stas Bekman, Amanpreet Singh, Anton Lozhkov, Thomas Wang, Siddharth Karamcheti, Alexander~M. Rush, Douwe Kiela, Matthieu Cord, and Victor Sanh.
\newblock Obelics: An open web-scale filtered dataset of interleaved image-text documents, 2023.

\bibitem[Lewis et~al.(2020)Lewis, Perez, Piktus, Petroni, Karpukhin, Goyal, K\"{u}ttler, Lewis, Yih, Rockt\"{a}schel, Riedel, and Kiela]{NEURIPS2020_6b493230_rag}
Patrick Lewis, Ethan Perez, Aleksandra Piktus, Fabio Petroni, Vladimir Karpukhin, Naman Goyal, Heinrich K\"{u}ttler, Mike Lewis, Wen-tau Yih, Tim Rockt\"{a}schel, Sebastian Riedel, and Douwe Kiela.
\newblock Retrieval-augmented generation for knowledge-intensive nlp tasks.
\newblock In H.~Larochelle, M.~Ranzato, R.~Hadsell, M.F. Balcan, and H.~Lin (eds.), \emph{Advances in Neural Information Processing Systems}, volume~33, pp.\  9459--9474. Curran Associates, Inc., 2020.
\newblock URL \url{https://proceedings.neurips.cc/paper_files/paper/2020/file/6b493230205f780e1bc26945df7481e5-Paper.pdf}.

\bibitem[Li et~al.(2023)Li, Zhang, Chen, Wang, Yang, and Liu]{li2023ottermultimodalmodelincontext}
Bo~Li, Yuanhan Zhang, Liangyu Chen, Jinghao Wang, Jingkang Yang, and Ziwei Liu.
\newblock Otter: A multi-modal model with in-context instruction tuning, 2023.
\newblock URL \url{https://arxiv.org/abs/2305.03726}.

\bibitem[Li et~al.(2024)Li, Zhang, Guo, Zhang, Li, Zhang, Zhang, Li, Liu, and Li]{li2024llavaonevisioneasyvisualtask}
Bo~Li, Yuanhan Zhang, Dong Guo, Renrui Zhang, Feng Li, Hao Zhang, Kaichen Zhang, Yanwei Li, Ziwei Liu, and Chunyuan Li.
\newblock Llava-onevision: Easy visual task transfer, 2024.
\newblock URL \url{https://arxiv.org/abs/2408.03326}.

\bibitem[Liu et~al.(2023{\natexlab{a}})Liu, Li, Wu, and Lee]{liu2023visualinstructiontuning}
Haotian Liu, Chunyuan Li, Qingyang Wu, and Yong~Jae Lee.
\newblock Visual instruction tuning, 2023{\natexlab{a}}.
\newblock URL \url{https://arxiv.org/abs/2304.08485}.

\bibitem[Liu et~al.(2023{\natexlab{b}})Liu, Li, Wu, and Lee]{liu2023visualllava}
Haotian Liu, Chunyuan Li, Qingyang Wu, and Yong~Jae Lee.
\newblock Visual instruction tuning.
\newblock In \emph{Thirty-seventh Conference on Neural Information Processing Systems}, 2023{\natexlab{b}}.
\newblock URL \url{https://openreview.net/forum?id=w0H2xGHlkw}.

\bibitem[Liu et~al.(2024)Liu, Song, Lin, Lam, Neubig, Li, and Yue]{liu2024visualwebbench}
Junpeng Liu, Yifan Song, Bill~Yuchen Lin, Wai Lam, Graham Neubig, Yuanzhi Li, and Xiang Yue.
\newblock Visualwebbench: How far have multimodal {LLM}s evolved in web page understanding and grounding?
\newblock In \emph{First Conference on Language Modeling}, 2024.
\newblock URL \url{https://openreview.net/forum?id=egVSgtJJAx}.

\bibitem[Liu et~al.(2023{\natexlab{c}})Liu, Iter, Xu, Wang, Xu, and Zhu]{liu-etal-2023-g-eval}
Yang Liu, Dan Iter, Yichong Xu, Shuohang Wang, Ruochen Xu, and Chenguang Zhu.
\newblock {G}-eval: {NLG} evaluation using gpt-4 with better human alignment.
\newblock In Houda Bouamor, Juan Pino, and Kalika Bali (eds.), \emph{Proceedings of the 2023 Conference on Empirical Methods in Natural Language Processing}, pp.\  2511--2522, Singapore, December 2023{\natexlab{c}}. Association for Computational Linguistics.
\newblock \doi{10.18653/v1/2023.emnlp-main.153}.
\newblock URL \url{https://aclanthology.org/2023.emnlp-main.153}.

\bibitem[Ma et~al.(2024)Ma, Zang, Chen, Chen, Jiao, Li, Lu, Liu, Ma, wen Dong, Zhang, Pan, Jiang, Wang, Cao, and Sun]{Ma2024MMLongBenchDocBL}
Yubo Ma, Yuhang Zang, Liangyu Chen, Meiqi Chen, Yizhu Jiao, Xinze Li, Xinyuan Lu, Ziyu Liu, Yan Ma, Xiao wen Dong, Pan Zhang, Liangming Pan, Yu-Gang Jiang, Jiaqi Wang, Yixin Cao, and Aixin Sun.
\newblock Mmlongbench-doc: Benchmarking long-context document understanding with visualizations.
\newblock \emph{ArXiv}, abs/2407.01523, 2024.
\newblock URL \url{https://api.semanticscholar.org/CorpusID:270870089}.

\bibitem[Masry et~al.(2022)Masry, Do, Tan, Joty, and Hoque]{masry-etal-2022-chartqa}
Ahmed Masry, Xuan~Long Do, Jia~Qing Tan, Shafiq Joty, and Enamul Hoque.
\newblock {C}hart{QA}: A benchmark for question answering about charts with visual and logical reasoning.
\newblock In Smaranda Muresan, Preslav Nakov, and Aline Villavicencio (eds.), \emph{Findings of the Association for Computational Linguistics: ACL 2022}, pp.\  2263--2279, Dublin, Ireland, May 2022. Association for Computational Linguistics.
\newblock \doi{10.18653/v1/2022.findings-acl.177}.
\newblock URL \url{https://aclanthology.org/2022.findings-acl.177}.

\bibitem[Mathew et~al.(2020)Mathew, Karatzas, Manmatha, and Jawahar]{Mathew2020DocVQAAD}
Minesh Mathew, Dimosthenis Karatzas, R.~Manmatha, and C.~V. Jawahar.
\newblock Docvqa: A dataset for vqa on document images.
\newblock \emph{2021 IEEE Winter Conference on Applications of Computer Vision (WACV)}, pp.\  2199--2208, 2020.
\newblock URL \url{https://api.semanticscholar.org/CorpusID:220280200}.

\bibitem[OpenAI(2023)]{gpt4vision}
OpenAI.
\newblock Gpt-4v(ision) system card, 2023.
\newblock URL \url{https://api.semanticscholar.org/CorpusID:263218031}.

\bibitem[Pfitzmann et~al.(2022)Pfitzmann, Auer, Dolfi, Nassar, and Staar]{doclaynet-10.1145/3534678.3539043}
Birgit Pfitzmann, Christoph Auer, Michele Dolfi, Ahmed~S. Nassar, and Peter Staar.
\newblock Doclaynet: A large human-annotated dataset for document-layout segmentation.
\newblock In \emph{Proceedings of the 28th ACM SIGKDD Conference on Knowledge Discovery and Data Mining}, KDD '22, pp.\  3743–3751, New York, NY, USA, 2022. Association for Computing Machinery.
\newblock ISBN 9781450393850.
\newblock \doi{10.1145/3534678.3539043}.
\newblock URL \url{https://doi.org/10.1145/3534678.3539043}.

\bibitem[Reddy et~al.(2024)Reddy, Koncel-Kedziorski, Lai, and Tanner]{Reddy2024DocFinQAAL}
Varshini Reddy, Rik Koncel-Kedziorski, Viet~Dac Lai, and Chris Tanner.
\newblock Docfinqa: A long-context financial reasoning dataset.
\newblock In \emph{Proceedings of ACL}, 2024.
\newblock URL \url{https://api.semanticscholar.org/CorpusID:266999305}.

\bibitem[Robertson \& Zaragoza(2009)Robertson and Zaragoza]{Robertson2009ThePRbm25}
Stephen~E. Robertson and Hugo Zaragoza.
\newblock The probabilistic relevance framework: Bm25 and beyond.
\newblock \emph{Found. Trends Inf. Retr.}, 3:\penalty0 333--389, 2009.
\newblock URL \url{https://api.semanticscholar.org/CorpusID:207178704}.

\bibitem[Saad-Falcon et~al.(2023)Saad-Falcon, Barrow, Siu, Nenkova, Rossi, and Dernoncourt]{saad2023pdftriage}
Jon Saad-Falcon, Joe Barrow, Alexa Siu, Ani Nenkova, Ryan~A Rossi, and Franck Dernoncourt.
\newblock Pdftriage: question answering over long, structured documents.
\newblock \emph{arXiv preprint arXiv:2309.08872}, 2023.
\newblock URL \url{https://arxiv.org/pdf/2309.08872}.

\bibitem[Shi et~al.(2023)Shi, Chen, Misra, Scales, Dohan, Chi, Sch\"{a}rli, and Zhou]{irrelevant-context-10.5555/3618408.3619699}
Freda Shi, Xinyun Chen, Kanishka Misra, Nathan Scales, David Dohan, Ed~Chi, Nathanael Sch\"{a}rli, and Denny Zhou.
\newblock Large language models can be easily distracted by irrelevant context.
\newblock In \emph{Proceedings of the 40th International Conference on Machine Learning}, ICML'23. JMLR.org, 2023.

\bibitem[Tanaka et~al.(2023)Tanaka, Nishida, Nishida, Hasegawa, Saito, and Saito]{Tanaka_Nishida_Nishida_Hasegawa_Saito_Saito_2023_slidevqa}
Ryota Tanaka, Kyosuke Nishida, Kosuke Nishida, Taku Hasegawa, Itsumi Saito, and Kuniko Saito.
\newblock Slidevqa: A dataset for document visual question answering on multiple images.
\newblock \emph{Proceedings of the AAAI Conference on Artificial Intelligence}, 37\penalty0 (11):\penalty0 13636--13645, Jun. 2023.
\newblock \doi{10.1609/aaai.v37i11.26598}.
\newblock URL \url{https://ojs.aaai.org/index.php/AAAI/article/view/26598}.

\bibitem[Verga et~al.(2024)Verga, Hofstatter, Althammer, Su, Piktus, Arkhangorodsky, Xu, White, and Lewis]{verga2024replacingjudgesjuriesevaluating}
Pat Verga, Sebastian Hofstatter, Sophia Althammer, Yixuan Su, Aleksandra Piktus, Arkady Arkhangorodsky, Minjie Xu, Naomi White, and Patrick Lewis.
\newblock Replacing judges with juries: Evaluating llm generations with a panel of diverse models, 2024.
\newblock URL \url{https://arxiv.org/abs/2404.18796}.

\bibitem[Wang et~al.(2024)Wang, Bai, Tan, Wang, Fan, Bai, Chen, Liu, Wang, Ge, Fan, Dang, Du, Ren, Men, Liu, Zhou, Zhou, and Lin]{wang2024qwen2vlenhancingvisionlanguagemodels}
Peng Wang, Shuai Bai, Sinan Tan, Shijie Wang, Zhihao Fan, Jinze Bai, Keqin Chen, Xuejing Liu, Jialin Wang, Wenbin Ge, Yang Fan, Kai Dang, Mengfei Du, Xuancheng Ren, Rui Men, Dayiheng Liu, Chang Zhou, Jingren Zhou, and Junyang Lin.
\newblock Qwen2-vl: Enhancing vision-language model's perception of the world at any resolution, 2024.
\newblock URL \url{https://arxiv.org/abs/2409.12191}.

\bibitem[Wang et~al.(2023)Wang, Lv, Yu, Hong, Qi, Wang, Ji, Yang, Zhao, Song, et~al.]{wang2023cogvlm}
Weihan Wang, Qingsong Lv, Wenmeng Yu, Wenyi Hong, Ji~Qi, Yan Wang, Junhui Ji, Zhuoyi Yang, Lei Zhao, Xixuan Song, et~al.
\newblock Cogvlm: Visual expert for pretrained language models.
\newblock \emph{arXiv preprint arXiv:2311.03079}, 2023.

\bibitem[Xiao et~al.(2024)Xiao, Mastrapas, and Wang]{xiao2024jina}
Han Xiao, Georgios Mastrapas, and Bo~Wang.
\newblock Jina {CLIP}: Your {CLIP} model is also your text retriever.
\newblock In \emph{Multi-modal Foundation Model meets Embodied AI Workshop @ ICML2024}, 2024.
\newblock URL \url{https://openreview.net/forum?id=lSDkG98goM}.

\bibitem[Yue et~al.(2023)Yue, Ni, Zhang, Zheng, Liu, Zhang, Stevens, Jiang, Ren, Sun, Wei, Yu, Yuan, Sun, Yin, Zheng, Yang, Liu, Huang, Sun, Su, and Chen]{yue2023mmmu}
Xiang Yue, Yuansheng Ni, Kai Zhang, Tianyu Zheng, Ruoqi Liu, Ge~Zhang, Samuel Stevens, Dongfu Jiang, Weiming Ren, Yuxuan Sun, Cong Wei, Botao Yu, Ruibin Yuan, Renliang Sun, Ming Yin, Boyuan Zheng, Zhenzhu Yang, Yibo Liu, Wenhao Huang, Huan Sun, Yu~Su, and Wenhu Chen.
\newblock Mmmu: A massive multi-discipline multimodal understanding and reasoning benchmark for expert agi.
\newblock \emph{arXiv preprint arXiv:2311.16502}, 2023.

\bibitem[Zhang et~al.(2024)Zhang, Patil, Jain, Shen, Zaharia, Stoica, and Gonzalez]{zhang2024raft}
Tianjun Zhang, Shishir~G Patil, Naman Jain, Sheng Shen, Matei Zaharia, Ion Stoica, and Joseph~E. Gonzalez.
\newblock {RAFT}: Adapting language model to domain specific {RAG}.
\newblock In \emph{First Conference on Language Modeling}, 2024.
\newblock URL \url{https://openreview.net/forum?id=rzQGHXNReU}.

\bibitem[Zhao et~al.(2024)Zhao, Zhang, Chia, Zhao, and Bing]{zhao2024autoarenallmsautomating}
Ruochen Zhao, Wenxuan Zhang, Yew~Ken Chia, Deli Zhao, and Lidong Bing.
\newblock Auto arena of llms: Automating llm evaluations with agent peer-battles and committee discussions, 2024.
\newblock URL \url{https://arxiv.org/abs/2405.20267}.

\bibitem[Zheng et~al.(2023)Zheng, Chiang, Sheng, Zhuang, Wu, Zhuang, Lin, Li, Li, Xing, Zhang, Gonzalez, and Stoica]{mt-judge-NEURIPS2023_91f18a12}
Lianmin Zheng, Wei-Lin Chiang, Ying Sheng, Siyuan Zhuang, Zhanghao Wu, Yonghao Zhuang, Zi~Lin, Zhuohan Li, Dacheng Li, Eric Xing, Hao Zhang, Joseph~E Gonzalez, and Ion Stoica.
\newblock Judging llm-as-a-judge with mt-bench and chatbot arena.
\newblock In A.~Oh, T.~Naumann, A.~Globerson, K.~Saenko, M.~Hardt, and S.~Levine (eds.), \emph{Advances in Neural Information Processing Systems}, volume~36, pp.\  46595--46623. Curran Associates, Inc., 2023.
\newblock URL \url{https://proceedings.neurips.cc/paper_files/paper/2023/file/91f18a1287b398d378ef22505bf41832-Paper-Datasets_and_Benchmarks.pdf}.

\bibitem[Zhu et~al.(2024)Zhu, Liu, Feng, Wang, Li, and Chua]{zhu2024tat}
Fengbin Zhu, Ziyang Liu, Fuli Feng, Chao Wang, Moxin Li, and Tat-Seng Chua.
\newblock Tat-llm: A specialized language model for discrete reasoning over tabular and textual data.
\newblock \emph{arXiv preprint arXiv:2401.13223}, 2024.

\bibitem[Zhu et~al.(2020)Zhu, Ahuja, Juan, Wei, and Reddy]{zhu2020question}
Ming Zhu, Aman Ahuja, Da-Cheng Juan, Wei Wei, and Chandan~K Reddy.
\newblock Question answering with long multiple-span answers.
\newblock In \emph{Findings of the Association for Computational Linguistics: EMNLP 2020}, pp.\  3840--3849, 2020.
\newblock URL \url{https://aclanthology.org/2022.naacl-main.391.pdf}.

\end{thebibliography}
\bibliographystyle{iclr2025_conference}

\appendix
\section{Appendix}

\subsection{Data Verification}
\label{sec:annotation}

To verify each question in our data construction process, we use the following guide to prompt the question generator models for automated verification.
Similarly, we use the same guide for human annotation in the human verification stage.

Based on the document content and question, answer yes or no only to the following questions:

1. Does the content contain any \{category\}?
2. Does the question require information from the \{category\}?
3. Is the question clear and answerable based on the \{category\}?
4. Is the question of reasonable difficulty and answer cannot be simply copied?

Where \{category\} refers to table or figure or text, which is denoted with the question.

Note:
If questions require general knowledge or commonsense in addition to the content, it is still acceptable.
In the document PDF file, each question is shown with the ID corresponding to excel sheet, and the document page as image
In the excel sheet, indicate “yes” or “no” for each check.

\subsection{Retrieval Methods}
\label{sec:retrieval}

To support our retrieval-based document question answering setting, we currently include four state-of-the art methods to retrieve relevant pages based on each question. They include text-based sparse methods such as BM25 \citep{Robertson2009ThePRbm25} embedding-based methods such as BGE-M3 \citep{chen2024bgem3embeddingmultilingualmultifunctionality}, multimodal piece-wise embedding methods such as JINA-CLIP \citep{xiao2024jina}, and multimodal page-wise embedding methods such as ColPali \citep{Faysse2024ColPaliED}. 
Note that piece-wise embedding methods separate encode each piece of text, table image, or figure image, whereas page-wise methods can encode the entire page content as a single image.
Thus, we rank each page in the document based on the similarity score or relevance score of that page with respect to the given question.
As each page may have multiple pieces of content, we consider the highest score of all pieces in a page to be the page-wise relevance score.
To compare the effectiveness of each method, we implement a standardized MRR score which refers to the mean reciprocal rank of the gold evidence page for each question. 
Based on the results in Table \ref{tab:retriever_comparison}, we find that ColPali which encodes each page as single image shows the best performance. 
Thus, we select ColPali as the preferred retrieval method in our main experiments.

\begin{table}[ht]
\centering
\begin{tabular}{lcccc}
\toprule
Retriever & Text & Figure & Table & All \\
\midrule
BM25 & 56.2 & 31.2 & 42.0 & 43.1 \\
CLIP & 57.1 & 37.9 & 50.4 & 48.5 \\
BGE-M3 & 66.4 & 36.4 & 53.6 & 52.1 \\
ColPali & 68.7 & 67.5 & 65.9 & 67.4 \\
\bottomrule
\end{tabular}
\caption{Retriever performance comparison.}
\label{tab:retriever_comparison}
\end{table}

\subsection{Evaluation Guide}
\label{sec:evaluation_guide}

To evaluate each model answer, we use the following scoring guide. Similarly, we use the same guide for human annotation in our analysis.

You will be given one response to a question based on a multimodal document containing texts, figures, or tables. Your task is to rate the response on correctness using a 1-5 scale. Please read and understand these instructions carefully, and keep them open for reference while reviewing.

Correctness (1-5) refers to how accurately, comprehensively, and appropriately the response addresses the question based on the information in the document.

5 - Fully Correct: Completely accurate, comprehensive, fully integrates relevant information from all parts of the document, and provides a coherent answer.

4 - Mostly Correct: Largely accurate with only minor errors or omissions, addresses most main points, and integrates information well.

3 - Partially Correct: Contains a mix of accurate and inaccurate information, addresses some key points but misses others, and partially integrates information.

2 - Mostly Incorrect: Has multiple inaccuracies, addresses only a small portion correctly, and shows minimal integration of information.

1 - Completely Incorrect: Contains significant errors, is irrelevant, or fails to address the question based on the document.

Evaluation Steps:
1. Thoroughly review the multimodal document and question.
2. Carefully read the response, comparing it to the document information.
3. Assess the response's accuracy, comprehensiveness, and relevance.
4. Assign a correctness score from 1 to 5 based on the criteria.

Question: {question}
Response: {answer}

Evaluation Form (score only without explanation)
Correctness:

\subsection{More examples}\label{sec:more_examples}
\subsubsection{Example of \datasetname{}}
Table~\ref{tab:two-chart-example} illustrates an example of a challenging question in our \datasetname{} benchmark. This question tests the ability of the model to identify and analyze trends across different charts and draw meaningful comparisons. 

\subsubsection{Case study of Retrieval-aware Tuning}
Table~\ref{tab:retrieve_aware_tuning_better_example} displays a sample question in \datasetname{} and the answers generated by Qwen2-VL and Qwen2-VL w/ Retrieval-aware Tuning. 
The answer generated by Qwen2-VL states that the Cosine method consistently shows the highest latent cosine similarity across all datasets, which is incorrect. In fact, the zero-shot stitching experiment does not involve the Cosine method. It appears that Qwen2-VL may have been misled by the keyword "cosine" appearing elsewhere in the retrieved context. 
In contrast, the answer generated by Qwen2-VL w/ Retrieval-aware Tuning correctly identifies that the affine method consistently obtains the highest latent cosine similarity (lcos) across all datasets. This demonstrates the effectiveness of our Retrieval-aware Tuning method in improving the model's capability to comprehend retrieved context.


\begin{table}[ht]
\centering
\begin{tabular}{p{0.95\textwidth}}
\toprule
    \textbf{Question:} \\
    How does the relationship between reference length percentile and the percentage of empty modes differ from the relationship between reference sentence length percentile and the probability of empty context? Explain the key differences in the trends shown by these two graphs. \\
    \midrule
    \textbf{Relevant page (truncated):} \\
    \begin{center}
    \includegraphics[width=0.85\textwidth]{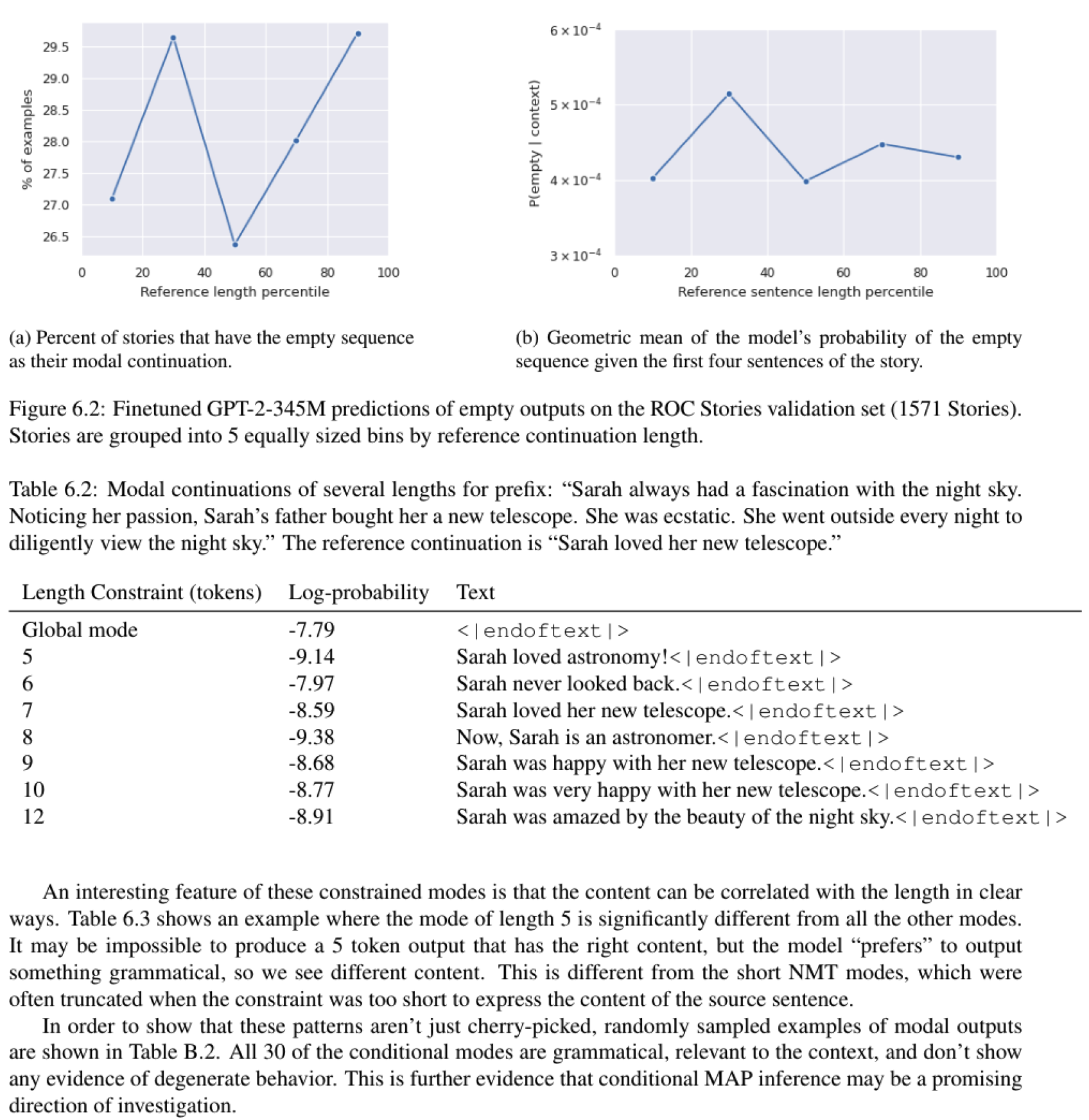}
    \end{center}\\
    \bottomrule
\end{tabular}
\caption{An example of a challenging question from \datasetname{} that requires the model to compare the trends of two charts in a document.}
\label{tab:two-chart-example}
\end{table}

\begin{table}[ht]
\centering
\small
\begin{tabular}{p{0.95\textwidth}}
\toprule
    \textbf{Question:} \\
    Which method consistently shows the highest latent cosine similarity (lcos)
across all datasets (MNIST, F-MNIST, CIFAR-10, CIFAR-100) in the zero-shot stitching
experiment, and what might this imply about the nature of transformations between latent spaces in
autoencoders? \\
    \midrule
    \textbf{Answer generated by Qwen2-VL:} \\
    The method that consistently shows the highest latent cosine similarity (lcos)
across all datasets (MNIST, F-MNIST, CIFAR-10, CIFAR-100) in the zero-shot stitching experiment
is the "Cosine" method. This implies that the "Cosine" method is able to capture the most similar
latent spaces across different architectures, modalities, and datasets. This suggests that the nature
of transformations between latent spaces in autoencoders is not exclusively correlated by
orthogonal transformations, but rather by a class of transformations larger than orthogonal
transformations. This finding highlights the importance of understanding and modeling the particular
class of transformations that control the correlation between NNs during image autoencoding. \\
\midrule
    \textbf{Answer generated by Qwen2-VL w/ Retrieval-aware Tuning:} \\
    The method that consistently shows the highest latent cosine similarity (lcos)
across all datasets (MNIST, F-MNIST, CIFAR-10, CIFAR-100) in the zero-shot stitching experiment
is the "affine" method. This method achieves the highest lcos values for all datasets, with scores of
0.94 for MNIST, 0.94 for F-MNIST, 0.96 for CIFAR-10, and 0.96 for CIFAR-100. This consistency in
high lcos values suggests that the affine transformation is particularly effective in aligning the latent
spaces of different autoencoders. The high lcos indicates a strong alignment between the latent
representations of the autoencoders, which implies that the affine transformation is capable of
capturing the underlying structure and relationships between the latent spaces more accurately
than other methods. The results imply that the nature of transformations between latent spaces in
autoencoders is not exclusively controlled by orthogonal transformations. The affine transformation,
which includes scaling, rotation, and translation, appears to be a more versatile and effective
method for aligning these spaces. This finding suggests that the latent spaces of image
autoencoders are related by a broader class of transformations, including those that are not strictly
orthogonal, which is a significant insight into the nature of these transformations and their
implications for model stitching and zero-shot learning. \\
    \midrule
    \textbf{Relevant page (truncated):} \\
    \begin{center}
    \includegraphics[width=0.6\textwidth]{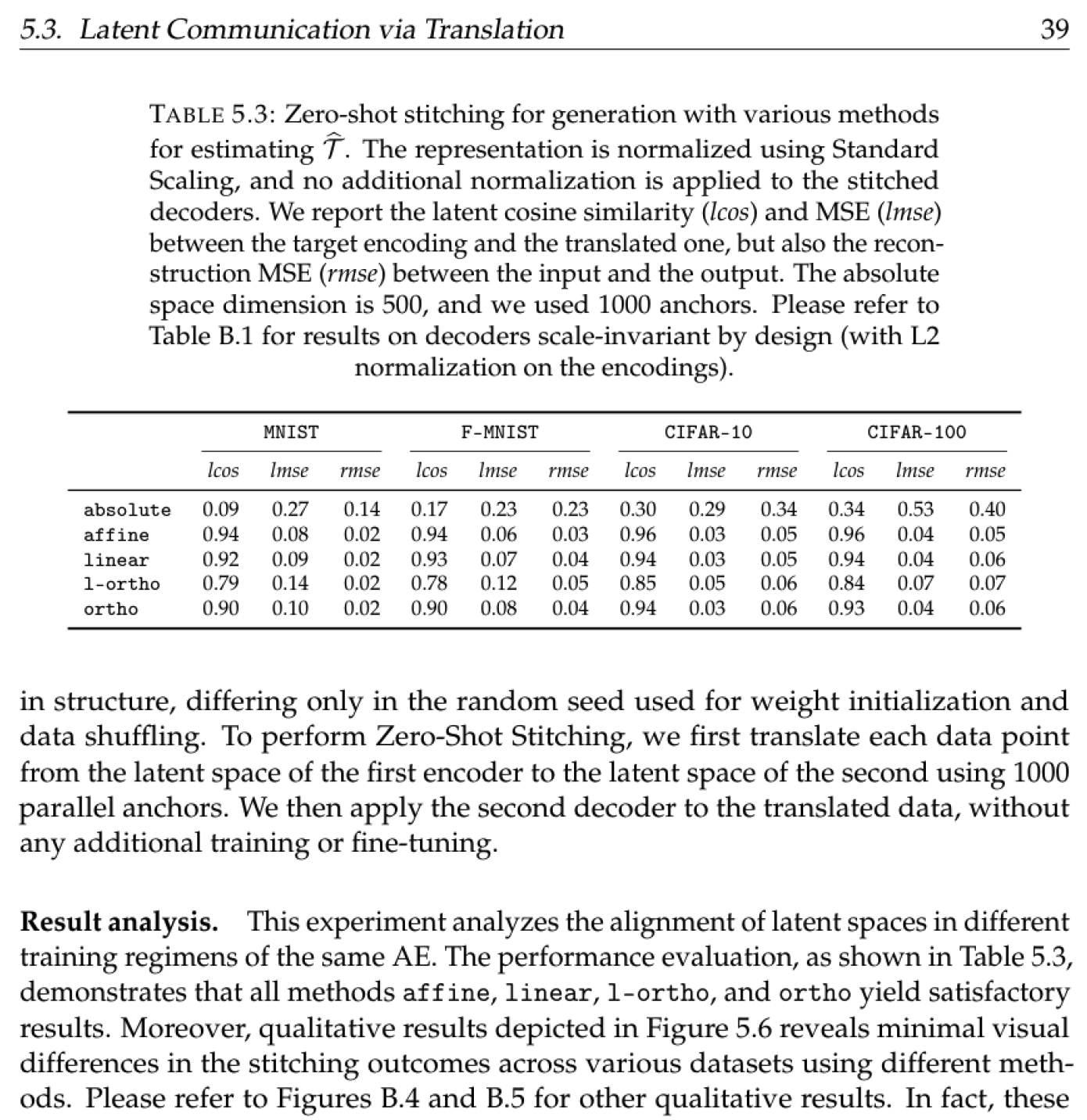}
    \end{center}\\
    \bottomrule
\end{tabular}
\caption{Sample answers generated by Qwen2-VL and Qwen2-VL w/ Retrieval-aware Tuning, respectively.}
\label{tab:retrieve_aware_tuning_better_example}
\end{table}

\end{document}